\pdfoutput=1

\documentclass[11pt]{article}

\usepackage[preprint]{acl}

\usepackage{times}
\usepackage{latexsym}
\usepackage{listings}
\usepackage{multirow}
\usepackage{booktabs}

\usepackage{colortbl}
\usepackage{etoolbox}
\usepackage{tabularx}
\usepackage{subcaption}
\usepackage[T1]{fontenc}

\usepackage[utf8]{inputenc}

\usepackage{microtype}

\usepackage{graphicx}

\newcommand{\hlnew}[1]{\textcolor{black}{#1}}

\title{Evaluating the Bias in LLMs for Surveying Opinion and Decision Making in Healthcare}

\author{
  Yonchanok Khaokaew$^1$  \\
  {y.khaokaew} \\
  {@unsw.edu.au}  \\\And
    Flora Salim$^1$  \\
  {flora.salim} \\
  {@unsw.edu.au}  \\\And
  Andreas Z\"ufle$^2$  \\
  {azufle@emory.edu} \\\And
    Hao Xue$^1$  \\
  {hao.xue1} \\
  {@unsw.edu.au}  \\\AND
  Taylor Anderson$^3$  \\
  {tander6@gmu.edu} \\\And
  C. Raina MacIntyre$^1$  \\
  {r.macintyre} \\
  {@unsw.edu.au}\\\And
  Matthew Scotch$^4$  \\
  {  matthew.scotch@asu.edu} \\\And
  David Heslop$^1$  \\
  {d.heslop} \\
  {@unsw.edu.au} \\\AND
  $^1$ University of New South Wales, Australia, $^2$ Emory University, USA\\
   {\bf$^3$ George Mason University, USA, $^4$ Arizona State University, USA}
  }

\begin{document}
\maketitle
\begin{abstract}
Generative agents have been increasingly used to simulate human behaviour in silico, driven by large language models (LLMs). These simulacra serve as sandboxes for studying human behaviour without compromising privacy or safety. However, it remains unclear whether such agents can truly represent real individuals. This work compares survey data from the Understanding America Study (UAS) on healthcare decision-making with simulated responses from generative agents. Using demographic-based prompt engineering, we create digital twins of survey respondents and analyse how well different LLMs reproduce real-world behaviours. Our findings show that some LLMs fail to reflect realistic decision-making, such as predicting universal vaccine acceptance. However, Llama 3 captures variations across race and Income more accurately but also introduces biases not present in the UAS data. This study highlights the potential of generative agents for behavioural research while underscoring the risks of bias from both LLMS and the prompting strategy.
\end{abstract}

\section{Introduction}

The rise of large language models (LLMs) has enabled advances in agentic artificial intelligence (AI), where AI systems can make independent choices and act autonomously~\cite{acharya2025agentic,park2023generative,xi2025rise,shanahan2023role}. Generative agents, in particular, have been shown to create realistic synthetic human populations, or simulacra, where individual agents follow daily life patterns and interact with each other~\cite{park2023generative,shanahan2023role,han2024ibsen, wang2024incharacter}. These simulacra offer a promising approach to studying human behaviour in silico, raising whether they can effectively model complex decision-making in real-world scenarios. In healthcare, where decisions are shaped by personal, social, and policy factors, the ability of simulacra to approximate human choices has significant implications. If LLMs can reliably simulate decision-making, they could serve as valuable tools for policy analysis, health behaviour prediction, and intervention design. However, their accuracy and potential biases when applied to real-world data require careful evaluation, particularly as LLMs have been shown to amplify racial biases in healthcare applications~\cite{ferrara2024butterfly}.

A key challenge in using LLMs for healthcare decision modelling is determining whether they effectively replicate factors shaping real-world health decisions. Unlike clinical diagnosis, which follows medical guidelines, social, economic, and behavioural influences shape choices such as seeking treatment or vaccination. While surveys provide structured insights into human intentions, LLMs offer a scalable alternative for modelling decisions in agent-based simulations. However, their ability to generate realistic health choices remains uncertain.

To investigate this, we compare health decision-making in a disease simulation framework, focusing on vaccination as a case study. Individuals make choices based on varying levels of contextual information, including personal risk perception, demographics, and external messaging. We compare LLM-generated vaccine decisions to survey data from the Understanding America Study (UAS)~\cite{kapteyn2024understanding}, which includes socioeconomic, risk perception, and personal belief data. This enables the assessment of LLMs' alignment with human decision patterns and potential biases diverging from real-world behaviours.

Despite this potential, several challenges remain. First, while LLMs generate human-like responses, it is unclear whether they truly capture the reasoning behind health-related decisions. Prior research suggests LLMs can retrieve medical knowledge, but their ability to simulate human decision processes is still in question~\cite{hager2024evaluation}. Since vaccination intentions are shaped by social and psychological factors, it is critical to assess whether LLMs accurately model these influences or merely reflect statistical patterns from their training data.

LLM-generated decisions may exhibit biases that diverge from human decision-making, raising concerns about their reliability in public health modelling. Biases in LLM training data can create demographic disparities~\cite{kim2025assessing}, making assessing them against actual human decisions essential. This study explores: \textbf{RQ1:} Can LLMs effectively model healthcare decisions, such as vaccination intentions?; \textbf{RQ2:} What biases emerge in LLM-generated decisions across demographic groups, and how do models distribute decisions among populations?

We hypothesise that LLMs can approximate human decision-making, but their effectiveness depends on the amount and type of contextual information provided (H1). Additionally, pretraining data and prompt formulation may cause LLMs to exhibit biases that differ from human biases (H2).

This study tests these hypotheses through a structured experiment. We compare LLM-generated vaccination decisions with survey responses to assess alignment and examine biases by analysing disparities across demographic groups. Our findings enhance an understanding of LLMs' strengths and limitations in modelling healthcare behaviours and decision-making.

\section{Method}

\begin{figure}
        \centering
        \includegraphics[width=0.92\linewidth]{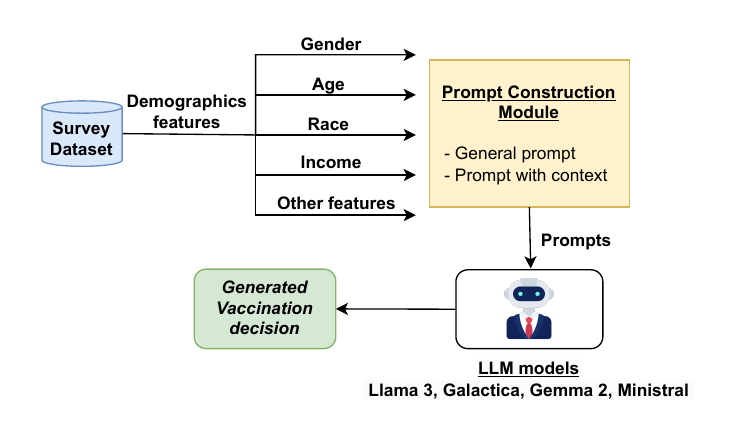}
        \caption{Overview of the experimental setup }
        \label{fig:ov}
\vspace{-15px}
\end{figure}

To analyse how LLMs approximate human decision-making in healthcare, we design a study that integrates demographic attributes, contextual prompts, and LLM-generated decisions (Figure \ref{fig:ov}). LLMs are prompted with structured demographic profiles under various pandemic scenarios, and their responses are analysed to assess decision patterns and potential biases.

We evaluate vaccination decisions by testing models across four historical pandemic contexts from 2020. Each model is presented with a standardised decision-making prompt, incorporating demographic details and situational factors. Model predictions are then compared to UAS survey data to assess alignment with real-world trends across different pandemic phases. To analyse biases, we examine disparities in LLM-generated decisions within each demographic category (e.g., variations in vaccine acceptance across racial or Income groups). \hlnew{To understand potential biases, we analyse both internal inconsistencies across demographic groups and compare model outputs to real-world survey data (UAS) where available. This helps identify whether LLMs reflect known behavioural differences or exaggerate group-level patterns.} This approach assesses LLMs' reliability in healthcare decisions and highlights potential biases from demographic variations in responses.

\subsection{Dataset}
Our study utilises data from the Understanding America Study’s Coronavirus in America survey~\cite{kapteyn2024understanding}, which tracks U.S. attitudes, health behaviours, and policy responses to COVID-19. We analyse data from the national long-form questionnaire, focusing on survey waves from March 2020 to January 2021 — a period when vaccine information was limited. \hlnew{While other surveys, such as the Household Pulse Survey (HPS) \cite{beleche2021covid}, provide valuable data, their vaccine-related responses only began in 2021, after broader vaccine availability and public awareness. We chose the UAS dataset because it uniquely covers the full timeline from the beginning of the pandemic, and also provides rich demographic, behavioural, and psychological features relevant to our simulation goals (see Appendix~\ref{app:UASfeature}).}

\subsection{Experimental Design}
\subsubsection{ Experimental Setup}
To investigate whether LLMs can approximate human decision-making in healthcare, we test LLMs on the question: \textit{"How likely are you to get vaccinated for coronavirus once a vaccination is available to the public?"}. Each model is prompted with demographic attributes (age, gender, Income, race, education, and worry level) to simulate individual decision-making. The responses are compared to real-world survey data from the UAS to evaluate alignment and detect biases in predictions.

To assess whether LLM-generated decisions reflect changes in public sentiment, we structure the experiment around four historical pandemic contexts in 2020: \textit{Jan–Mar} (early outbreak, economic uncertainty, healthcare preparations), \textit{Apr–Jun} (lockdowns, financial hardship, overwhelmed hospitals), \textit{Jul–Sep} (reopening, second-wave concerns, vaccine trials), and \textit{Oct–Dec} (U.S. election, emergency vaccine approval, economic relief).

We assess each LLM to see how contextual variations influence decision-making. We also analyse bias to identify disparities in LLM responses regarding vaccine acceptance and whether demographic details mitigate biases. Each model generates 11.5k samples across all demographic profiles and four pandemic phases. We apply majority voting over three generations per prompt to reduce response variance. \hlnew{In designing the prompt, we experimented with different prompt formats, including persona-only, location-based, situation-based, and few-shot prompts. However, few-shot prompts often made the model repeat or exaggerate specific examples, especially when sensitive attributes like race or Income were included. To reduce this risk, we chose a zero-shot prompt with contextual information, which gave more consistent and balanced results.}

\subsubsection{Model Selection and Specifications}
We evaluate four instruction-tuned open-source LLMs:  \textit{Meta Llama-3-8B-Instruct}~\cite{dubey2024llama}, optimized for instruction-following with reinforcement learning from human feedback (RLHF); \textit{Google Gemma-2-9B-IT}~\cite{team2024gemma}, designed for improved generalization and contextual understanding; \textit{Galactica-6.7B-Evol-Instruct}~\cite{taylor2022galactica}, fine-tuned for structured instruction-following and domain-specific knowledge; and \textit{Mistral AI Ministral-8B-Instruct}~\cite{jiang2023mistral}, known for balancing efficiency and reasoning performance. 

\hlnew{These models were selected for their architectural diversity, public availability, and relevance to our task. Galactica, in particular, has been used in recent clinical decision-making work with LLMs \cite{poulain2024bias}. We focused on open-source models to support reproducibility, especially for researchers who may not have access to commercial APIs or large-scale compute. While we tested GPT-3.5 in early experiments, full-scale comparisons were not feasible due to cost constraints.}

\subsubsection{Evaluation Metrics}

We employ two key metrics, the Disparate Impact Ratio and Jensen-Shannon Divergence, to assess bias and alignment in LLM-generated decisions.

\textit{Disparate Impact Ratio (DIR)} measures disparities in decision distributions across demographic groups~\cite{feldman2015certifying}. It is defined as: $DIR = \frac{\min(P_i)}{\max(P_i)}$, where \( P_i \) represents the probability of vaccine acceptance for each demographic category. When multiple categories exist, we compute the ratio of the best to worst outcomes to identify the largest disparity. A DIR near 1 suggests fair treatment, while lower values indicate significant disparities.

\textit{Jensen-Shannon Divergence (JSD)} quantifies differences in decision distributions \cite{lin1991divergence} within LLM-generated outputs across demographic groups (e.g., Male vs. Female, White vs. Black vs. Asian). A higher JSD value indicates greater inconsistencies in decision patterns, suggesting potential demographic biases in the model's decision-making.

\section{Result}

\begin{figure}
    \centering
+    \includegraphics[width=0.92\linewidth]{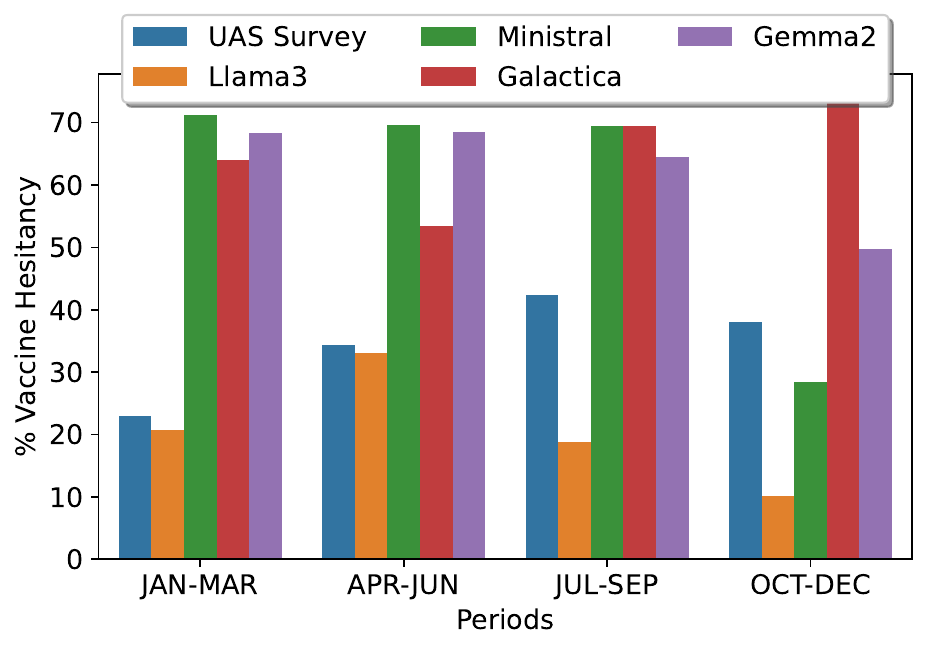}
    \caption{Comparison of survey and LLMs decision outputs 4 different situations}
    \label{fig:compare4situ}
\vspace{-9px}
\end{figure}
\subsection{Comparison of LLM predictions with UAS survey data}

Our study examines vaccination intentions using four LLMs, prompting them with pandemic-phase-specific contexts and demographic profiles. The structured prompts cover four COVID-19 phases in the U.S. (Jan–Dec 2020), allowing us to assess how LLMs simulate decision-making trends compared to UAS survey data (Figure \ref{fig:compare4situ}). \hlnew{vaccine acceptance is defined as the percentage of responses where the model (or survey participant) answered that they would not get vaccinated.}

In Jan–Mar 2020, early uncertainty led to a moderate hesitancy of ($\approx$25\%) in the UAS survey. Llama3 closely matched this, while Ministral and Galactica significantly overestimated. By Apr–Jun 2020, as lockdowns and economic strain intensified, hesitancy increased slightly. Llama3 remained the closest match, while Galactica and Gemma2 continued to overestimate, though their predictions adjusted slightly.

During Jul–Sep 2020, vaccine trials progressed, but concerns over a second wave grew. Hesitancy exceeded 40\%; Gemma2 aligned well, but Llama3 underestimated, and Galactica and Ministral still overpredicted. In Oct–Dec, with vaccine approvals, hesitancy remained under 40\%. Ministral matched best, while Llama3 again underestimated, and Galactica/Gemma2 overestimated.

\hlnew{These results reveal consistent behavioural tendencies across models. Llama3 appears to favour early optimism in vaccine uptake, possibly influenced by its alignment tuning and preference for socially desirable responses. In contrast, Galactica and Ministral maintain higher levels of skepticism throughout, potentially due to their training on domain-specific or scientific content that emphasises uncertainty and risk. These behavioural biases suggest that outputs are shaped not only by prompts, but also by training and alignment, highlighting the need to evaluate LLM bias when applying them to public health simulations.}

\subsection{Bias analysis in LLM-generated decisions}

\begin{table}[h]
\footnotesize
\caption{Disparate Impact Ratio (DIR) and Jensen-Shannon Divergence (JSD) across models.}
\begin{center}
\resizebox{0.7\textwidth}{!}{\begin{tabularx}{\textwidth}{c|c|c|c|c|c|c|c|c|}
 \cmidrule{1-9}
& \multicolumn{2}{|c|}{\textbf{Llama3}} & \multicolumn{2}{|c|}{\textbf{Gemma2}} & \multicolumn{2}{|c|}{\textbf{Ministral}}&\multicolumn{2}{|c|}{\textbf{Galactica}} \\
 \cmidrule{1-9}
\textbf{Feature Set} &DIR&JSD&DIR&JSD&DIR&JSD&DIR&JSD\\\cmidrule{1-9}
Gender& 0.918 & 0.004& 0.853 & 0.002& 0.933 & 0.000& 0.974 & 0.0001\\ 
Race& 0.942 & 0.001& 0.602 & 0.009& 0.864 & 0.001& 0.973 & 0.0000\\ 
Income& 0.624 & 0.019& 0.236 & 0.035& 0.404 & 0.013& 0.889 & 0.0001\\ 
Education& 0.614 & 0.066& 0.061 & 0.140& 0.349 & 0.045& 0.941 & 0.0020\\
\cmidrule{1-9}
\cmidrule{1-9}

\end{tabularx}}
\label{table:dirJSD}
\end{center}
\vspace{-5px}
\end{table}
\hlnew{We define bias in this context not as the presence of group-level differences, but as distortions or misalignment from real-world trends. Our goal is to examine whether LLM-generated disparities reflect known behavioural patterns or diverge in ways that may misrepresent specific groups. Our results reveal significant disparities in vaccine decisions generated by LLMs across income, education, and race, as shown in Table~\ref{table:dirJSD}. Gemma2 and Ministral exhibit the most pronounced disparities, with low DIR values for income (0.236 and 0.404) and education (0.061 and 0.349). Their high JSD scores also indicate substantial divergence across groups. Prior work has shown that income and education correlate with vaccine acceptance~\cite{aw2021covid, allen2021factors}, but the gap in model outputs suggests a disconnect between LLM behaviour and real-world trends.}

\begin{figure}
    \centering
    \includegraphics[width=0.9\linewidth]{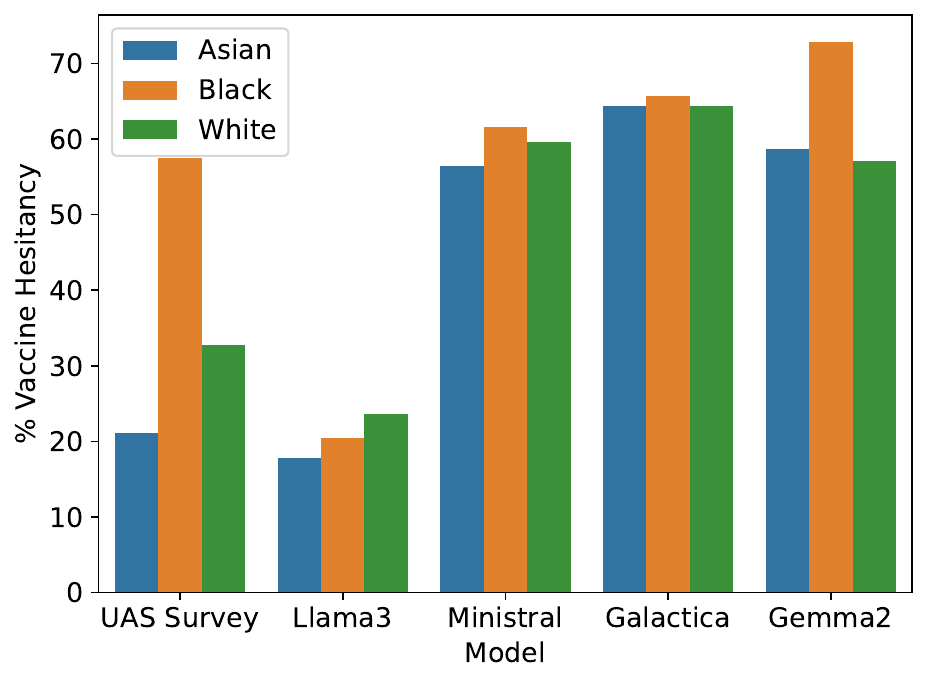}
    \caption{Racial Bias in Vaccine Decisions: LLM Outputs vs. Survey Data}
    \label{fig:rraceb}
\vspace{-10px}
\end{figure}

\hlnew{Galactica produces the most balanced predictions across demographic groups, while Llama3 performs moderately but tends to underestimate hesitancy overall. Among the demographic dimensions, we focus on race in Figure~\ref{fig:rraceb} due to the distinct trends observed across models. The rightmost bars in the figure represent UAS survey data, which shows higher vaccine acceptance among Black respondents compared to White and Asian groups. Most LLMs, however, flatten this distribution. For instance, Llama3 shows uniformly low hesitancy across races, and Galactica and Ministral generate nearly equal predictions for all groups. Gemma2 reflects the real-world ranking more closely and predicts the highest hesitancy for Black individuals—over 70\%—but still reduces the contrast compared to the UAS survey.}

\hlnew{These findings underscore the importance of bias-aware evaluation in LLM-driven simulations. Rather than amplifying disparities, most models in our study obscure them, failing to reproduce the degree of variation seen in survey data. This form of misalignment can still be problematic, as it may lead to overly neutral or inaccurate behavioural predictions. Future work should explore mitigation strategies such as refining prompts, improving demographic representation in training data, and adopting fairness-aware modelling techniques.}

\section{Conclusion}

\hlnew{Our findings suggest that LLMs show potential in simulating decision-making behaviors, but careful evaluation is needed to ensure alignment with real-world patterns. We observe that some models, like Llama3, closely follow early vaccine acceptance trends but underestimate skepticism in later phases, possibly due to their alignment tuning. Other models, such as Galactica and Ministral, display more cautious tendencies, potentially reflecting their training on domain-specific or scientific content emphasising uncertainty and risk.}

\hlnew{Importantly, we do not assume that all demographic groups should behave identically. Instead, we evaluate whether the models reflect known behavioural differences accurately or distort them. Comparisons with UAS survey data and internal consistency metrics reveal that some models amplify racial, income, or education disparities, underscoring the need for bias-aware design in using LLMs for simulation.}

\hlnew{Future work should include a broader range of datasets, deeper prompt sensitivity studies, and an extended set of models (e.g., commercial LLMs such as GPT-4 or Deepseek). This is a crucial step toward understanding how LLMs can serve as reliable components in behavioural simulations and policy modelling.}

\section*{Limitations}

We show that LLM-based simulacra of human individuals show the potential to be used as a surrogate model for real surveys on human populations, specifically with Llama3 and Galactica capturing the effects of gender, race, Income, and education in vaccine acceptance surveys quite well. One limitation of this study is that these LLMs may have been exposed to the UAS Survey Data used for evaluation. While the UAS data is not publicly available and should not have been used for model training, it is possible that the dataset — or scientific summaries of related sources such as the ACS — may have been indirectly incorporated during pretraining. Another limitation is the use of a single dataset (UAS) as the primary reference for model evaluation. We acknowledge that this limits generalisability. Although we considered other datasets such as the HPS survey, vaccine hesitancy data in HPS begins in 2021 — after broad vaccine availability and increased public awareness. Our study focuses on early-stage decision-making under limited information, which the UAS dataset uniquely captures, beginning in early 2020.

In addition, while the UAS panel is large (14,700 respondents in 2024, designed to represent the U.S. population), it is still subject to sampling variance. Another shortcoming of our work is using UAS as the ground truth to evaluate model bias — a necessary simplification in this initial study. Finally, we note that LLM-generated decisions may amplify real-world disparities rather than simply reflect them. This raises concerns about their reliability in behavioural modelling. Future research should explore mitigation strategies to reduce bias propagation in AI-driven decision-making and assess model robustness across more diverse datasets. A more systematic evaluation of prompt sensitivity and its effect on model outputs is planned for future work, including testing the impact of different prompt styles (e.g., few-shot, simplified persona prompts) across demographic groups.

\section*{Acknowledgements}
 This research is supported by the United States National Science Foundation under Grants No. 2302968, No. 2302969, and
 No. 2302970, titled "Collaborative Research: NSF-CSIRO: HCC: Small: Understanding Bias in AI Models for the Prediction of Infectious Disease Spread", as well as by the Australian Commonwealth Scientific and Industrial Research Organisation (CSIRO), and National Science Foundation under Grants No. 2125530 and No. 2041952. The project described in this paper relies on data from survey(s) administered by the Understanding America Study, which is maintained by the Center for Economic and Social Research (CESR) at the University of Southern California. So, research reported in this publication was partially supported by the National Institute on Aging of the National Institutes of Health and in part by the Social Security Administration under Award Number U01AG077280. The content is solely the responsibility of the authors and does not necessarily represent the official views of the National Institutes of Health. We express our gratitude to the NVIDIA Academic Grant Program for providing access to an A100 GPU on Saturn Cloud. We further thank OpenAI’s Researcher Access Program for API access to GPT models. (For GPU and API) 


\section*{Declaration of use of AI assistants}
This paper was developed with the assistance of generative AI tools (ChatGPT and Copilot) to improve clarity, structure, and conciseness. AI-assisted technologies were used for text refinement, grammar correction, and formatting but did not contribute to conceptualisation, analysis, or interpretation results. After using these tools, the author(s) reviewed and edited the content as needed and take(s) full responsibility for the content of the publication.
\bibliography{main}

\appendix

\section{Appendix}
\label{sec:appendix}

\subsection{Features in UAS dataset}\label{app:UASfeature}
The Understanding America Study (UAS) dataset features include demographic attributes and behavioural indicators relevant to vaccine decision-making. The demographic attributes consist of gender (Male/Female), age (15–99), and race (White, Black, Asian). Socioeconomic factors include household income, categorised into eight bins from `Less than \$25,000' to `\$200,000 and above', and education level, classified as "High school or less", "Some college", and "Bachelor or more". Psychological factors, such as worry levels over the past two weeks, are grouped into four categories: `Not at all', `Several days', `More than half the days', and `Nearly every day'. The survey also includes vaccination intent, which is recorded as a binary outcome (Yes/No).

For prompt generation, we selected representative values for each attribute: ages (18-99), genders (Male, Female), races (White, Black, Asian), and income levels spanning eight bins. Education was categorized into three levels, and worry levels followed the original survey classification. These structured inputs allowed us to systematically analyse how LLMs simulate vaccine decision-making across different demographic groups.




\subsection{Prompt template}

Imagine yourself in the following situation: [SITU PROMPT]. Your background and personal circumstances are as follows: [You are a {AGE}-year-old {GENDER} of {RACE} ethnicity, living in a diverse country with varying access to healthcare, differing levels of trust in government and medical institutions, and socioeconomic disparities. Your annual Income is {INCOME}. Your education level is {EDU\_LEVEL}. Over the past two weeks, you have been worrying about your health {WORRY\_LEVEL}]. Please use this persona to answer the question below: 

\textit{`How likely are you to get vaccinated for coronavirus once a vaccination is available to the public?' }

In this context, please answer based on your persona. Answer: [Yes/ No] Short reason: [FILL IN] based on your persona

The [SITU PROMPT] will be replaced by the following contextual prompt from a different period:

\textbf{January - March 2020}: From January to March 2020, COVID-19 emerged in the US, leading to the first reported cases and the declaration of a pandemic by the WHO. The early economic impact included business closures and rising unemployment while the healthcare system began preparing for an influx of patients. Consider the initial response to the virus, the economic impact, and healthcare system preparations.

\textbf{April - June 2020}: During April to June 2020, the US experienced strict lockdown measures, a surge in unemployment, and significant strain on the healthcare system due to COVID-19. Businesses were closed, and many people faced financial hardships. Healthcare workers were overwhelmed, and there were shortages of essential medical supplies. Considering these challenges and public health measures

\textbf{July - September 2020}: From July to September 2020, states in the US began to reopen, leading to mixed responses in terms of economic recovery and public health. Concerns about a second wave of COVID-19 emerged as cases began to rise again in some areas. Progress was made in vaccine development, with several candidates entering late-stage trials. Include considerations of reopening efforts, second-wave concerns, and progress in vaccine development.

\textbf{October - December 2020}: In the period from October to December 2020, the US presidential election took place, creating significant political and social implications. COVID-19 vaccines received emergency use authorization in December, leading to the beginning of vaccination campaigns. Additional economic relief measures were implemented to support individuals and businesses affected by the pandemic. 

\subsection{LLMs licensing, Data usage approval and Generation parameters}
All LLMs used in this study were accessed through Hugging Face, with the necessary licences acquired before the experiments. The generation parameters were configured with a temperature of 0.6 and a top-p of 0.9, which allowed for controlled randomness in responses while maintaining coherence. Additionally, approval for using the Understanding America Study (UAS) survey data was obtained in accordance with its usage policies.

\end{document}